\pdfoutput=1

\documentclass[11pt]{article}

\usepackage[]{ACL2023}

\usepackage{times}
\usepackage{latexsym}
\usepackage{hyperref}
\usepackage{enumitem}
\usepackage[T1]{fontenc}
\usepackage[utf8]{inputenc}
\usepackage{microtype}
\usepackage{inconsolata}

\usepackage{graphicx}
\usepackage{longtable}
\usepackage{comment}
\usepackage{multirow}
\usepackage{booktabs}

\usepackage{array}
\usepackage{ragged2e}
\newcolumntype{P}[1]{>{\RaggedRight\hspace{0pt}}p{#1}}

\usepackage{tabularx}
\usepackage{booktabs}
\DeclareGraphicsExtensions{.pdf,.ai,.jpg,.png}
\setkeys{Gin}{pagebox=artbox}

\usepackage{tabularx}
\usepackage{booktabs}

\title{Towards a Perspectivist Turn in Argument Quality Assessment}

\author{
Julia Romberg\textsuperscript{1}, 
Maximilian Maurer\textsuperscript{1,3}, 
Henning Wachsmuth\textsuperscript{2} \and 
Gabriella Lapesa\textsuperscript{1,3} \\
\textsuperscript{1}GESIS - Leibniz Institute for the Social Sciences \\ 
\textsuperscript{2}Leibniz University Hannover \\  
\textsuperscript{3}Heinrich-Heine University Düsseldorf\\
\textsuperscript{1}\texttt{first.last@gesis.org},
\textsuperscript{2}\texttt{h.wachsmuth@ai.uni-hannover.de}
}

\begin{document}
\maketitle

\begin{abstract}
The assessment of argument quality depends on well-established logical, rhetorical, and dialectical properties that are unavoidably subjective: multiple valid assessments may exist, there is no unequivocal ground truth. This aligns with recent paths in machine learning, which embrace the co-existence of different perspectives. However, this potential remains largely unexplored in NLP research on argument quality. One crucial reason seems to be the yet unexplored availability of suitable datasets. We fill this gap by conducting a systematic review of argument quality datasets. We  assign them to a multi-layered categorization  targeting two aspects: (a) What has been annotated: we collect the quality dimensions covered in datasets and consolidate them in an overarching taxonomy, increasing dataset comparability and interoperability. (b) Who annotated: we survey what information is given about annotators, enabling perspectivist research and grounding our recommendations for future actions. To this end, we discuss datasets suitable for developing perspectivist models (i.e., those containing individual, non-aggregated annotations), and we showcase the importance of a controlled selection of annotators in a pilot study.
\end{abstract}
\section{Introduction}
\label{sec:introduction}

The question of ``what makes an argument good'' is at the core of computational argumentation (CA), the area of natural language processing (NLP) dealing with the mining, assessment, and generation of arguments \cite{stede_argumentation_2019, lauscher2022survey}. While rooted in well-established theories, argument quality (AQ) still exhibits a high degree of subjectivity in perception. This degree may vary across quality aspects; for example, evaluating an argument's benefit in agreement-seeking discussions is considered less subjective than assessing its effectiveness \cite{wachsmuth-etal-2017-computational}.\,\,

The CA community is aware of the variance in annotators' perception \cite{stab-gurevych-2014-annotating, teruel-etal-2018-increasing, hautli-janisz-etal-2022-disagreement}. Not least, this is documented by the generally moderate inter-annotator agreement in AQ annotations --- a widely accepted condition that authors commonly attribute to the subjective nature of the task (e.g., \citealt{wachsmuth-etal-2017-computational, Gretz_Friedman_Cohen-Karlik_Toledo_Lahav_Aharonov_Slonim_2020, ng-etal-2020-creating, ziegenbein-etal-2023-modeling}).%
\footnote{Certainly, not all disagreement is due to subjectivity. We refer the reader to the Limitations section for a discussion.}
\citet{wachsmuth-werner-2020-intrinsic} explicitly question whether an aggregated ground truth is suitable to model AQ.

Meanwhile, the NLP community has started to undergo a fundamental change in the way it deals with subjective tasks. While aggregated ground truth and an according model alignment were long standard, more recent work calls for this course to be reconsidered \cite{basile2020end,plank-2022-problem,cabitza_toward_2023,Frenda2024}: Rather than eradicating any existence of annotator disagreement, the \emph{perspectivist turn} embraces the co-existence of perspectives \cite{uma-etal-2021-semeval, davani-etal-2022-dealing, leonardelli-etal-2023-semeval}. This transformation implies the acceptance of variations in data annotation (through non-aggregated datasets) as well as the consideration of heterogeneity in modeling and evaluation \cite{uma-2021-survey, basile-etal-2021-need, plank-2022-problem}.

We postulate that the perspectivist turn in NLP lends itself as a natural solution to face the issue of subjectivity in modeling AQ. Not only does it have a better shot at promoting diversity and fairness in AQ assessment, such as allowing for valid but minority voices \cite{noble2012minority, prabhakaran-etal-2021-releasing}. It is also likely to be more robust in modeling perceptions of AQ across (changing) societies (e.g., today’s minority groups may become tomorrow’s majority) and target audiences.

Yet, the perspectivist turn so far had only minimal impact on AQ. Presumably, one reason for the limited modeling of perspectives in AQ is the lack of datasets designed for this purpose. Preference has been given to aggregated annotations, whereas individual labeling decisions were often not communicated (e.g., \citealt{persing-ng-2013-modeling, park-cardie-2018-corpus, toledo-etal-2019-automatic, ijcai2022p575}). While a solution may be new datasets, the annotation of argumentation phenomena is highly complex and costly. We therefore deem it essential to first \textit{gain an overview of the options that existing datasets already offer for developing perspectivist models}. This is the goal of the paper at hand. 

We provide a systematic literature review of 103 AQ {datasets and their properties}.%
\footnote{The resulting database can be accessed publicly here: \href{https://github.com/juliaromberg/perspectivist-turn-aq}{https://github.com/juliaromberg/perspectivist-turn-aq}}
Crucially, to support the perspectivist turn, our collection includes meta-information about annotators and the availability of non-aggregated annotations. While only 24 datasets come with the latter, 14 of them seem relevant to the perspectivist turn. In a pilot study, we conduct a statistical analysis of the disagreement patterns in four of them. We conclude by highlighting the opportunities of available datasets and discuss challenges, for example a lack of transparency and socio-demographic diversity.

\paragraph{Contributions}
(1) We release an extensive database with 32 types of meta-information about 103 AQ datasets. (2) We review the multitude of annotated AQ categories 
(\textit{what is annotated}) and consolidate them in an overarching taxonomy to foster comparability and interoperability.
(3) We perform a comprehensive meta-analysis of annotators (\textit{who annotates}) across the datasets, uncovering a lack of transparency and socio-demographic diversity, promoting bias in AQ datasets and models.
(4) We deep-dive into the 24 datasets with non-aggregated labels both qualitatively and quantitatively, and discuss their potential for a perspectivist turn in AQ.

\section{Related Work}
\label{sec:relatedwork}

\paragraph{Surveys of Computational Argumentation}

In the last 20 years, the field of CA has witnessed a constant development driven by the potential for real world applications, but also by the increasingly interdisciplinary shape that the field has assumed. The number of  surveys on CA is a clear sign of this progress: ranging from foundational work that set up or updated the conceptual coordinates for the field \cite{peldszus2013argument,stede_argumentation_2019,lawrence2020survey,lauscher2022survey}, to surveys with a data-driven focus \cite{cabrio2018survey,SchaeferStede+2021+45+58}, to specific advances in NLP, such as generation \citep{wang2023survey}. 

Specifically for AQ, \citet{wachsmuth-etal-2017-computational} introduced a first holistic systematization of the field according to AQ dimensions. \citet{wachsmuth-etal-2024-argument} update the survey, taking into account the challenges and potentials for the employment of large language models (LLMs) in AQ assessment. While focused on applications of CA for social good as a whole, the survey by \citet{vecchi-etal-2021-towards} puts a strong interdisciplinary focus on AQ and its interface with deliberation quality.

\textit{No survey so far has targeted a systematic categorization of datasets}.%
\footnote{Dataset repositories for CA do exist, i.e., \href{https://github.com/acidrobin/arglu-repo}{\textit{ARGLU}} and the \href{https://webis.de/data.html\#other-corpora}{Webis database}, but are by no means comprehensive and lack our conceptual categorization and focus on annotators.}
This is the gap we fill:%
\footnote{While this paper was under review, a survey on AQ was published by \citet{ivanova-etal-2024-lets}, highlighting the timeliness of this topic. The authors examined the state of AQ research in general, whereas we focus on its future transition into a task where human label variation plays a significant role. Due to a different search strategy, our survey covers a broader range of datasets (103 compared to 32), and offers a more in-depth analysis (with 32 manually annotated meta-categories, compared to 10). Crucially, we adopt a timely interdisciplinary taxonomy that integrates argument and deliberation quality, which \citet{ivanova-etal-2024-lets} also hint at for future research.} 
we survey {datasets}, focusing on the {consolidation of covered AQ categories into an overarching taxonomy}, and who the {annotators} are.

\paragraph{Dimensions of Argument Quality}
\label{subsec:AQ}

\citet{wachsmuth-etal-2017-computational}'s taxonomy is the most commonly adopted one for AQ assessment. Rooted in argumentation theory, it emphasizes three aspects:

\textit{Logical cogency} and its subcategories promote a valid reasoning process at the level of individual arguments. An argument is considered cogent if its premises are rationally worthy of being believed to be true (\textit{local acceptability}), its premises contribute to the acceptance or rejection of its conclusion (\textit{local relevance}), and if they provide enough support to make the conclusion rational (\textit{local sufficiency}).

\textit{Rhetorical effectiveness} and its subcategories mirror the persuasive power of an author's argument towards a target audience. Characteristics are a clear style (\textit{clarity}), maintaining a tone appropriate to the issue (\textit{appropriateness}), presenting components of the argument in a proper order (\textit{arrangement}), establishing the author's credibility (\textit{credibility}), and evoking emotions that make the audience more receptive (\textit{emotional appeal}).

\textit{Dialectical reasonableness} and its subcategories evaluate the contribution to resolving differences of opinions on a discussion level. Argumentation is deemed reasonable if the consideration and presentation of the arguments put forward for the issue are acceptable to the target audience (\textit{global acceptability}), contribute to the issue's resolutions (\textit{global relevance}), and adequately rebut the contestable counterarguments (\textit{global sufficiency}).

\citet{vecchi-etal-2021-towards} proposed to include \textit{deliberative norms} as a further aspect of AQ. This dimension incorporates democratic values into the dialectical view, adherence to which is particularly relevant to political arguments, but also applies to broader contexts like online communication. While the authors resorted specifically to the \emph{Discourse Quality Index} \cite{steenbergen2003dqi}, communication science has come up with various instruments to empirically measure deliberation quality (e.g., \citealt{stromer-galley2011deliberation, black2011deliberation, graham2003deliberation}). 

The exact criteria of (good) deliberation and consequently the instruments for measuring it are matter of controversial discussion \cite{dellicarpini2003deliberation}. \citet{friess2015deliberation} identified seven dimensions that are prevalent across various frameworks: Deliberative discourse should be an exchange grounded in \textit{rationality}. The exchange should take place through listening, understanding and actively responding to each other's opinions in a substantive way (\textit{interactivity}). Furthermore, deliberation should foster \textit{equality} by equipping all sides with the same opportunity to participate in the discussion and \textit{civility} for a respectful interaction. Arguments should be oriented towards the \textit{common good} of the community, and \textit{constructive} in finding a consensus decision for the issue of discussion. The last dimension relates to the use of \textit{alternative forms of communication} (e.g., storytelling).

\paragraph{Perspectivism and Argument Quality}

AQ assessment is a prime example of a subjective task: beyond logical well-formedness (and even there), the question of good arguments is bound to be answered in conflicting ways by annotators with different features (e.g., socio-demographics, life experiences, personality, and values) \cite{lukin-etal-2017-argument,durmus-cardie-2019-corpus,el-baff-etal-2020-analyzing}. This makes AQ an ideal perspectivist topic. 

Yet, \textit{perspectivist AQ assessment is only at its beginning, also because of the need for suitable data}. As datasets will be reviewed in the remainder of the paper, we focus here on the few works that have specifically targeted the {modeling of annotator perspectives} in AQ, i.e., by integrating label variation in the machine learning workflow. The first explicit step was taken by \citet{romberg-2022-perspective}, who predicted the subjectivity of the annotation as an indicator for trustworthiness of majority vote models. What is more, \citet{heinisch-etal-2023-architectural} compared approaches for modeling annotator-specific behavior.
\section{Systematic Review of Datasets}
\label{sec:review}

\paragraph{Search Methodology}
We searched in all major publication organs that, according to our experience in the field, cover the topic of CA. We included the leading conferences in computational linguistics (the entire ACL anthology, including the  Argument Mining Workshop), artificial intelligence (all from AAAI.org, IJCAI, and ICAIL), information retrieval (SIGIR and ECIR), and the specialized computational argumentation series COMMA.

To pre-filter a set of candidates, we used all pairs of search terms from \textit{\{argument, argumentation, argumentative, debate, deliberation, deliberative\} $\times$ \{quality, strength, persuasiveness, fallacies\}}. The retrieval was carried out with the Google site search including all papers that had been published until August 20, 2024, resulting in 238 candidate papers of which we found 42 to be relevant. Additionally, we employed a snowballing method \cite{wohlin2014snowballing} to ensure that the field is covered as completely as possible: looking at studies that either cite one of the previously identified papers (forward snowballing, with Google Scholar) or are cited by those papers (backward snowballing) led to further 56 relevant papers. In total, we identify 98 relevant papers, distributed among research communities as follows: NLP (73), artificial intelligence (6), information retrieval (7), CA (1), and further venues from computer science (4) and the social sciences (7). Appendix \ref{sec:review-details} describes the process in detail.

\paragraph{Categorization Taxonomy}
We assess the relevance of datasets by drawing from the taxonomies of \citet{wachsmuth-etal-2017-computational} and \citet{friess2015deliberation}. A paper is considered relevant if it introduces a new dataset (or extends an existing one) that at least loosely matches one or more of the following categories (whose theoretical background has been introduced in Section \ref{subsec:AQ}): i) \textit{logical cogency} with subcategories \textit{local acceptability}, \textit{local relevance}, and \textit{local sufficiency}; ii) \textit{rhetorical effectiveness} with subcategories \textit{clarity}, \textit{appropriateness}, \textit{arrangement}, \textit{credibility}, and \textit{emotional appeal}; iii) \textit{dialectical reasonableness} with subcategories \textit{global acceptability}, \textit{global relevance}, and \textit{global sufficiency}; iv) \textit{deliberative norms} with subcategories \textit{rationality}, \textit{interactivity}, \textit{equality}, \textit{civility}, \textit{common good reference}, \textit{constructiveness}, and \textit{alternative forms of communication}; and \textit{overall argument quality}. Appendix \ref{app:taxonomy} provides the taxonomy and exact definitions in full.

\paragraph{Categorization Reliability}
Mapping of dataset dimensions according to the AQ taxonomy was conducted by the first author of the paper. To validate this process, 10 papers (\textasciitilde 10\%) were reassigned to two other authors. For the high-level categories, we reached Fleiss' $\kappa$ values of $1.0$ for logical cogency, $0.73$ for rhetorical effectiveness, $0.71$ for dialectical reasonableness, and $0.70$ for deliberative norms, demonstrating robust inter-annotator reliability. The mean agreement across all 23 categories was lower, $0.52$. It is worth pointing out, though, that one of the other authors reached $0.72$ with the first author, which is why we deem our categorization to be reasonably reliable. A clear source of disagreement arose from the categorization of fallacies into the taxonomy. Reasoning errors (i.e., fallacies) can affect all dimensions of AQ, and we refined the annotation guidelines accordingly.

\paragraph{Collection of Meta-information}
In addition to the AQ categories, we gathered further information about the datasets. This includes general details such as genre, modality, language, and the availability of the dataset and annotation guidelines. While dataset availability was generally good (84 public or upon request), information on annotation guidelines was less available (32). We contacted authors of datasets without clear indications, encouraging public release in line with open science principles. As a result, 14 additional datasets now have publicly accessible guidelines, for a total of 11 unique guidelines made available.

We also collected a variety of characteristics that are of interest when looking through the perspectivist glasses. Most notable for the paper at hand are meta-information about annotators and the availability of non-aggregated annotations. Appendix~\ref{sec:database-details} lists all information contained in the database.
\section{Datasets: Annotations \& Annotators}
\label{sec:analysis}

The 98 identified papers introduce 103 AQ datasets in total. A complete list including the mapped AQ categories is in Appendix \ref{sec:relevant-datasets}. In what follows, we provide an overview of the quantitative properties arising from the comparison of their \textit{annotation}, and we then focus on  \textit{annotator} meta-information.

\subsection{Annotations: What Argument Quality?}

\begin{figure}[t]
  \includegraphics[scale=1]{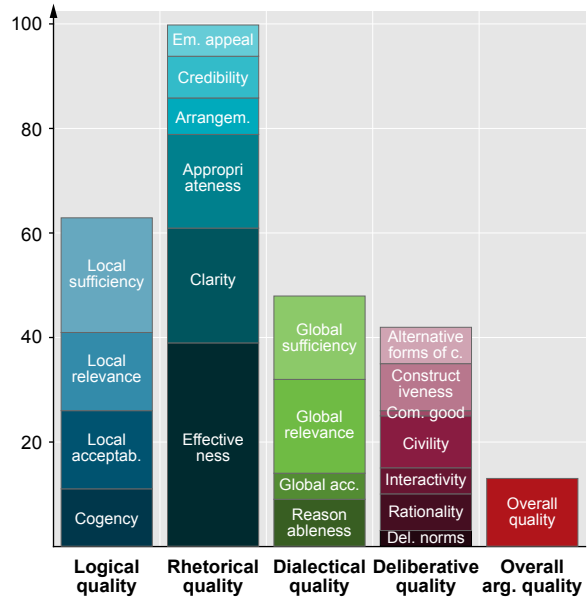}
  \caption{Frequency and distribution of AQ categories (major and sub-categories) as assigned to datasets, grouped by the four major categories and overall AQ.}
  \label{fig:overview-aq}
\end{figure}

Figure \ref{fig:overview-aq} shows the distribution of datasets among the categories of AQ. Particular interest can be observed for the rhetorical effectiveness of arguments, likely driven by its practical relevance and the availability of pre-annotated resources, such as the Reddit forum ChangeMyView and other online debate platforms where users rate the persuasiveness of each other's arguments. Increased attention has also been paid to the logical validity of arguments. However, the two dialectically-driven dimensions of reasonableness and deliberative norms received less attention, a finding that coincides with the CA community's constant call for a greater focus on dialogical argumentation \cite{ruiz-dolz-etal-2024-overview}.

Looking into the individual sub-categories of AQ, we find that almost all of them are covered. The only exception is \emph{equality}. However, we acknowledge that measuring whether participants have equal opportunities in a deliberation is challenging, as it extends beyond simply assessing active participation to the potential for participation depending on the socioeconomic capital that the participants hold \cite{friess2015deliberation}.

\begin{table}[t]
\centering
\small\renewcommand{\arraystretch}{1.2}
\setlength{\tabcolsep}{2pt}
\scriptsize
\begin{tabular}{P{2cm}p{5.4cm}}
\toprule
\bf Meta-Categories & \bf Specifications (Counts, Sorted Descending)\\
\midrule
Genre	&  social media (20), online debate portal	(19), persuasive essays	(12), crowd-sourced	(10), public participation (8), news articles (7), political debate	(6), web (5), collaborative online discussions (5), news comments (4), reviews (3), educational debate (2), fact-checking portals (2), QA forums (2), e-mail communication (2), Wikipedia (2), online educational material (1), classroom discussions (1), business model pitches (1), LLM-sourced (1) \\
Modality & text (100), multimodal (3)\\
Language & en (89), de (13), fr (4), jp (2), es (1), it (1), nl (1), pt-br (1), zh (1)\\
\midrule
Manual annotation & manual (82), automatic (18), manual+automatic (3)\\
Selection method of annotators & students/available	(22), consistency with experts (18), expertise (16), language competence (16), reliability checks (11), performance in prior tasks (8), educational level (5), balanced sample wrt. to some property (5), consistency with fellow annotators (3)\\
In-house annotators or crowd-workers & in-house (39: experts 13, novice 11, mixed 5, n.a. 10), crowd (27; task expertise unknown), in-house+crowd (5: in-house experts 4, in-house mixed 1), n.a. (14)\\
Annotator attributes (across in-house and crowd-sourcing  datasets) & no indication at all (45); education (25), age (18), native language (14), gender (11), profession (9), professional background (8), stance (5), country of origin (4), country of residence (3), occupation (3), political view (3), nationality (2), personality traits (2), annotation time (1), civic engagement (1), competence (1), employment status (1), ethnicity (1), income (1), race (1), religion (1),  role (1), spirituality (1)\\
\bottomrule
\end{tabular}
\caption{Counts of specifications for different meta-categories on datasets (top) and annotations (bottom).}
\label{tab:dataset-stats}
\end{table}

The upper part of Table \ref{tab:dataset-stats} provides an overview of selected dataset properties. In terms of genre, the datasets cover a great variety, with social media, online debate portals, and persuasive essays being the most prominent. Seven datasets draw from multiple genres \cite{xu2014identifying, napoles-etal-2017-finding, lauscher-etal-2020-rhetoric, ziegenbein-etal-2023-modeling, falk-lapesa-2023-storyarg, helwe-etal-2024-mafalda, li-etal-2024-side}. While most of the datasets focus solely on text, three are multimodal \cite{liu-etal-2022-imagearg, liu-etal-2023-overview, mancini-etal-2024-multimodal}. With respect to languages, we observe a very imbalanced situation with English accounting for over 85\% of the datasets. Four of the datasets contain multiple languages \cite{Gerber_Baechtiger_Shikano_Reber_Rohr_2018, toledo-ronen-etal-2020-multilingual, falk-lapesa-2022-scaling, reveilhac2023comparing}.

\subsection{Annotators: Whose Perspectives?}
\label{sec:perceptions}

We now take a closer look at the individuals that provide the AQ assessments, in order to understand whose perspectives current datasets cover. Table~\ref{tab:dataset-stats} shows the statistics on manual annotation, annotator selection and attributes. The majority of datasets were created through coordinated manual data annotation; fewer than 20\% of datasets were generated automatically by parsing existing internet resources, with the ground truth labels derived from a natural sample of platform users.

In manual annotation, authors indicated a variety of reasons for the selection of annotators, among them predominantly consistency with experts, expertise of the annotators themselves, and language competence. Students were also a common choice (in one quarter of the datasets), which might be an indicator for selection upon availability. Only five datasets had annotators selected with the aim of weighting socio-demographic characteristics according to certain standards, such as the representation of a country's population \cite{lukin-etal-2017-argument, brenneis-etal-2021-will}, balancing political ideologies or gender \cite{el-baff-etal-2018-challenge, falk-etal-2024-moderation}, and annotators from diverse debating circuits  \cite{joshi-etal-2023-arganalysis35k}. Also noteworthy is that in three datasets, annotators were excluded if inconsistent with fellow annotators' label decisions.

Looking more closely at the socio-demographic background of AQ annotators, we find that authors only occasionally provide information (in the papers or datasets). In case of \textit{in-house annotators}, we find a higher education in all cases indicated and often a background in NLP and related fields. A key differentiator is expertise in AQ, which separates in-house annotators into two groups: \textit{experts} and \textit{novice} annotators (usually students). Additionally, we find seven explicit mentions of gender (two datasets include both binary genders without specifying proportions, two use balanced samples, two have significantly more male annotators, and one includes two female and one male annotator). Age was reported twice, with ranges of 18--53 and 18--22 years. In case of \textit{annotators recruited on crowd-sourcing platforms}, socio-demographic information is reported sparsely, only for 8 datasets. In these cases, it is either used to draw a more representative sample or serves to narrow down the selection of annotators to the language of data.

On a more general note, characteristics that may invoke some bias in assessment such as political views (and related stances) were rarely collected, and there is similarly little information on cultural diversity among annotators. Individual characteristics that go beyond socio-demographic features are hardly at issue, except from \citet{lukin-etal-2017-argument} and \citet{el-baff-etal-2018-challenge} who collect personality traits, recognizing the potential impact on AQ perception.

\section{Towards Perspectivist Argument Quality Assessment}
\label{section-towards_perspectivist-quality-assessment}

\begin{table*}[t]
\centering
\scriptsize
\renewcommand{\arraystretch}{0.90}
\setlength{\tabcolsep}{2.5pt}
\begin{tabular}{p{0.31\linewidth}@{}rrrrp{0.43\linewidth}}
\toprule
\bf Dataset & \bf Size & \bf Per-item & \bf Total & \bf Category & \bf Annotators' Attributes Provided in Dataset\\ 
\midrule
\multicolumn{6}{c}{\em Introduced as non-aggregated to facilitate perspectivist machine learning or to promote diversity in annotations}\\
\midrule
CrowDEA Ideas \cite{Baba_Li_Kashima_2020} & 16,000 & 20 & 257 & crowd & - \\
Argument Concreteness \cite{romberg-etal-2022-corpus} & 1,127 & 5 & 5 & novice & - \\
TYPIC \cite{naito-etal-2022-typic} & 197 & 1--2 & 4 & in-house & - \\
Argument Validity Novelty \cite{heinisch-etal-2023-architectural} & 1,474 & 3 & 5 & expert & - \\
MAFALDA \cite{helwe-etal-2024-mafalda} & 268 & 1--4 & 4 & expert & - \\
UMOD \cite{falk-etal-2024-moderation} & 1,000 & 9 & 90 & crowd & race, gender, age, annotation time, role, competence, stance\\
\midrule
\multicolumn{6}{c}{\em Built to explore how argument perception differs between groups and individuals}\\
\midrule
Persuasion \& Personality \cite{lukin-etal-2017-argument} & 100 & 20 & 637 & crowd & personality traits, age, gender, political view, education, civic engagement, religion, spirituality, employment status, income, stance \\
Webis-Editorial-Quality-18 \cite{el-baff-etal-2018-challenge} & 1,000 & 6 & 24 & crowd & political view, personality traits \\
\midrule
\multicolumn{6}{c}{\em Personalization}\\
\midrule
n.a. \cite{hunter2017empirical} & 30 & 50 & 50 & crowd & - \\
SIGIR-19 \cite{potthast2019argument} & 494 & 1 & 40 & in-house & age, gender, stance \\
argumentation-attitude \cite{brenneis-etal-2021-will} & 946 & 1--147 & 674 & crowd & stance \\
\midrule
\multicolumn{6}{c}{\em Aggregated ground truth datasets that were released together with the individual labeling decisions}\\
\midrule
Dagstuhl-ArgQuality \cite{wachsmuth-etal-2017-computational} & 320 & 3 & 3 & expert & - \\
n.a. \cite{wachsmuth-etal-2017-argumentation} & 320 & 10 & 102 & crowd & - \\
n.a. \cite{mirzakhmedova2024reliable} & 320 & $\leq$ 10 & 108 & novice & - \\
GAQCorpus \cite{lauscher-etal-2020-rhetoric} & 5,285 & 1--13 & 27 & exp, crowd & - \\
EuropolisAQ \cite{falk-lapesa-2022-scaling} & 513 & 1--2 & 2 & expert & - \\
ArgQ! \citet{silva2021quality} & 352 & 4 & 4 & expert & - \\
UKPConvArg1 \cite{habernal-gurevych-2016-argument}  & 16,000 & 5 & 3,900 & crowd & stance \\
UKPConvArg2 \cite{habernal-gurevych-2016-makes} & 70,000 & 5 & 776 & crowd & - \\
Essay Argument Organization \cite{persing-etal-2010-modeling} & 1,003 & 1-6 & 6 & novice & - \\
Appropriateness Corpus \cite{ziegenbein-etal-2023-modeling} & 2,191 & 3 & 3 & crowd & - \\
UKP-InsufficientArgs \cite{stab-gurevych-2017-recognizing} & 433 & 3 & 3 & expert & - \\
Webis-ArgRank-17 \cite{wachsmuth-etal-2017-pagerank} & 110 & 7 & 7 & expert & - \\
StoryARG \cite{falk-lapesa-2023-storyarg} & 2,451 & 1--4 & 4 & in-house & - \\
\bottomrule
\end{tabular}
\caption{Overview of AQ datasets that come with non-aggregated annotations. In each case, we provide annotator counts \emph{per-item} and \emph{total}, \emph{categorize} them as in-house (experts, novice, or in-house; if expertise is unspecified) or crowd workers, and specify the \emph{annotators' attributes} contained directly in the datasets at the individual level.}
\label{tab:dataset_sizes}
\end{table*}

Developing perspectivist models requires the existence of multiple assessment perspectives. Among the 103 datasets we found, only 24 come with non-aggregated annotations. We detail these datasets, before we exemplify the potential impact of annotator groups on AQ assessment.

\subsection{Non-Aggregated Datasets}
\label{sec:non-aggr-datasets}

Table \ref{tab:dataset_sizes} lists the datasets with properties relevant to perspectivist model development. We identify four conceptual blocks: Six datasets were exclusively introduced as {non-aggregated for perspectivist approaches or to promote annotation diversity}. Two were developed to study {how argument perception varies based on group-level or individual characteristics}. Three stem from {personalization in argument retrieval}, and 13 are {aggregated datasets released together with the individual labeling decisions}. An extensive description of all 24 datasets is provided in Appendix~\ref{sec:description-non-aggr-datasets}. Here, we focus on those that we deem most useful for the perspectivist turn.

\paragraph{Populations}

For developing well-generalizable models, it is integral that the datasets represent a specific population, whose composition of perspectives can be learned. The annotations of four datasets were collected in a controlled setup in this regard: The \textit{Persuasion \& Personality} corpus \cite{lukin-etal-2017-argument} was created to study differences in the perception of argument effectiveness. Stance changes elicited by social media arguments were recorded from a representative sample of the US population. The \textit{argumentation-attitude} dataset \cite{brenneis-etal-2021-will} covers personalized views of strong arguments from a political opinion platform, rated by a representative sample of the German online population, in terms of age, gender, and education. \textit{Webis-Editorial-Quality-18} \cite{el-baff-etal-2018-challenge} captures differing perceptions of effectiveness in US news editorials on a balanced sample of liberals and conservatives, \textit{UMOD} \cite{falk-etal-2024-moderation} annotates characteristics of user-driven online moderation (including comment constructiveness), using a gender-balanced population.

Crowd-platform annotations sourced from a sufficiently large group of workers may also be assumed to approximate the broader population from the respective platform to a certain extent: \textit{UKPConvArg1} and \textit{UKPConvArg2} \cite{habernal-gurevych-2016-makes,habernal-gurevych-2016-argument} capture argument convincingness and the AQ reasons behind, with 16k and 70k items, respectively, and over 4k crowd annotators from the US. \textit{CrowDEA Ideas} \cite{Baba_Li_Kashima_2020} contains preference labels of 257 crowd workers for 16k solutions to an issue. These three datasets also stand out in their size; they are the only datasets with tens of thousands of annotated items.

Lastly, we highlight two datasets that bring together different groups of annotators for the same data. The \textit{Dagstuhl-ArgQuality} dataset \cite{wachsmuth-etal-2017-computational} of online debate arguments was rated by experts, novice student annotators (with no prior experience in CA) \cite{mirzakhmedova2024reliable},%
\footnote{The official release includes only a subset of annotators, but the authors kindly provided the full set upon request.} 
and crowd workers \cite{wachsmuth-etal-2017-argumentation} across the 15 dimensions of their taxonomy. The \textit{GAQCorpus} \cite{lauscher-etal-2020-rhetoric}, which includes diverse arguments annotated for cogency, effectiveness, reasonableness, and overall AQ, was annotated by a mix of 3 expert annotators and 24 crowd workers. Some items were annotated only by the crowd, others by the experts, and part of the dataset was jointly annotated by both groups. The different groups can be regarded as different types of populations and thus represent an interesting testing ground for group-specific analysis.

\paragraph{Individuals}

To model the perspectives of individuals more accurately, meta-information about them is needed. Such attributes can also help to build perspectivist models at the group level. The Persuasion \& Personality corpus, the argument-attitude dataset, WEBIS-Editorial-Quality-18, UMOD, and UKPConvArg1 provide several relevant attributes, including socio-demographics, stances on certain topics, and personality traits. In addition, the \textit{SIGIR-19} dataset \cite{potthast2019argument}, which codes logical, rhetorical, and dialectical AQ, includes gender, age, and stance for each annotator.

Besides, two datasets from our overview have already been used successfully in modeling human label variation, the non-aggregated version of the \textit{Argument Validity and Novelty} dataset \cite{heinisch-etal-2023-architectural} and the \textit{Argument Concreteness} corpus \cite{romberg-etal-2022-corpus}. Both lack background information on the annotators, but this limitation was initially secondary to the goal of developing perspectivist models for personalization \cite{heinisch-etal-2023-architectural}.

\subsection{Potential Impact of Annotator Groups}
\label{sec:pilot}

In Section \ref{sec:analysis}, we found that in-house annotators form a relatively homogeneous group concerning education and work background, with expertise being a key differentiator between experts and novices. In contrast, annotators recruited via crowdsourcing platforms can be assumed to represent a much more diverse sample in terms of socio-demographic attributes and lived experiences. We thus study two research questions on the impact of annotator groups on AQ annotation and prediction:
\begin{enumerate}[leftmargin=1cm]
\setlength{\itemsep}{0pt}
\item[RQ1.]
How comparable are annotations across annotator groups, how stable within them?
\item[RQ2.] 
How does this impact the performance bounds of models trained on group-specific annotations when transferred across groups?
\end{enumerate}

\paragraph{Data}
We use the two mentioned non-aggregated datasets with a mix of annotator groups: Dagstuhl-ArgQuality \citep{wachsmuth-etal-2017-computational} and its extensions \cite{mirzakhmedova2024reliable, wachsmuth-etal-2017-argumentation} (summarized as \textit{Dagstuhl}; annotated by experts, novice student annotators and crowd-platform workers), and the GAQCorpus \cite{lauscher-etal-2020-rhetoric} (\textit{GAQ}; annotated by experts and crowd-platform workers). For Dagstuhl, we resort to the 304 arguments deemed argumentative in the original corpus. For GAQ, we use the 538 arguments annotated by both experts and crowd annotators. We focus on cogency, effectiveness, reasonableness, and overall AQ as categories.

\paragraph{Experimental Setup}
For RQ1, we compute inter-annotator agreement (IAA) within and across annotator groups, using Krippendorff's~$\alpha$. For a full picture, we include all annotators per group.

To answer RQ2, we assume a situation in which a \textit{perfect model}, trained on the annotations of one group, is evaluated on another group, effectively performing a {population transfer}. We opted for simulation rather than real training of a model in order to minimize confounding factors, such as model deficits due to limited training data. This way, we can clearly illustrate the discrepancy that arises when population characteristics, and the differences in perspectives they encode, are ignored. We investigate the actual upper performance bounds in two scenarios, a traditional \textit{aggregated approach} with a single regression output per argument, and a \textit{perspectivist approach} in which we assume to obtain a learned regression label distribution per argument as the system output.

\begin{table}[t]
\centering
\scriptsize
\renewcommand{\arraystretch}{1}
\setlength{\tabcolsep}{4.2pt}
\begin{tabular}{p{0.5cm}lrrrr}
\toprule
&\bf Group  & \bf Cogency & \bf Effectiveness & \bf Reasonableness & \bf Overall\\
\midrule
\multirow{3}{*}{\parbox{0.6cm}{\textbf{Dag-\\stuhl}}} 
& E & \bf .372 & \bf .314 & \bf .437 & \bf .443 \\
& N & .230 & .208 & .197 & .233 \\
& C & .099 & .107 & .111 & .140 \\
& E, N & .114 & .098 & .134 & .126 \\
& E, C & .129 & .121 & .143 & .180 \\
& N, C& .060 & .072 & .071 & .083 \\
& E, N, C & .083 & .085 & .098 & .115 \\
\midrule
\textbf{GAQ} 
& E & \bf .175 & \bf .272 & \bf .258 & \bf .254 \\
& C & .156 & .148 & .154 & .173 \\
& E, C & .142 & .142 & .150 & .165 \\
\bottomrule
\end{tabular}
\caption{Krippendorff's $\alpha$ for different groups of annotators; experts (E), novice (N), and crowd workers (C).}
\label{tab:iaa}
\end{table}

\begin{table}[t]
\centering
\scriptsize
\renewcommand{\arraystretch}{1}
\setlength{\tabcolsep}{3.1pt}
\begin{tabular}{p{0.5cm}lrrrrrrrr}
\toprule
&& \multicolumn{2}{c}{\bf Cogency} & \multicolumn{2}{c}{\bf  Effectiven.} & \multicolumn{2}{c}{\bf  Reasonab.} & \multicolumn{2}{c}{\bf  Overall} \\ 
\cmidrule(lr){3-4}\cmidrule(lr){5-6}\cmidrule(lr){7-8}\cmidrule(lr){9-10}
&\bf  Transfer& \bf MAE &\bf WS& \bf MAE& \bf WS &\bf MAE& \bf WS &\bf MAE & \bf WS \\
\midrule
\multirow{3}{*}{\parbox{0.6cm}{\textbf{Dag-\\stuhl}}} &E, N & .697&  .159 & .714 & .155& .605& .161 & .686 & .158\\
&E, C & .499&.195 & .530&.194 & .463&.184 & .430&.179 \\
&N, C & .507&.164 &.470&.169 & .454&.162 & .480&.163 \\
\midrule
\textbf{GAQ} & E, C &.697& .111&.751& .122& .629& .110 &  .659&\ .109\\
\bottomrule
\end{tabular}
\caption{Group transfer evaluation for the aggregated approach (MAE) and the perspectivist approach (mean WS); experts (E), novice (N), and crowd workers (C).}
\label{tab:cross-group-dist}
\end{table}

\begin{figure*}[t]
    \centering
    \includegraphics[scale=1]{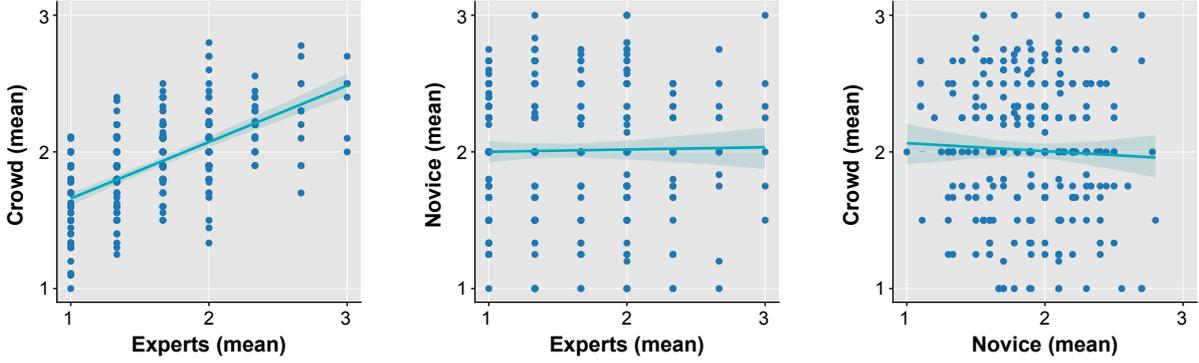}
    \caption{Instance-based aggregation of label decisions for overall AQ, assessed on a scale from 1 (low) to 3 (high), between two annotator groups on Dagstuhl, with a fitted linear regression model highlighting their relationship.}
    \label{fig:oq-aggregated}
\end{figure*}

To compare label distributions (i.e., the perfectly predicted one and that of a target population), we calculate the Wasserstein distance (WS) between the label distributions per item. We report the mean across the whole dataset, respectively. For aggregated regression outputs (i.e., mean ratings), we calculate the mean average error (MAE). We report results per dataset and quality dimension.

\paragraph{Results}
Regarding RQ1, Table \ref{tab:iaa} shows the IAA for each annotator group and their different combinations for both corpora. For in-group, across corpora and quality dimensions, we find that expert annotators have the highest agreement, though still comparably low. For Dagstuhl, crowd annotators have the lowest IAA. Across groups, IAAs drop compared to the involved group with the highest agreement for all combinations of groups and both corpora. This indicates varied annotations within groups and high disagreement across groups.

Regarding RQ2, Table \ref{tab:cross-group-dist} shows the results for both the aggregated and the perspectivist approach. For the former, we find that even if a model perfectly learns one group's aggregated annotation behavior, the minimum MAE achievable, in the worst case, is $0.714$ for Dagstuhl and $0.751$ for GAQ (both effectiveness). Figure \ref{fig:oq-aggregated} exemplifies what this means on Dagstuhl for predictions of overall AQ: while transfer from experts to crowd annotators and vice versa retains the same general trend, the other combinations behave effectively at random. This is in line with IAA per group combination.

The perspectivist evaluation, using label distributions, shows that the highest Wasserstein distances occur for cases of group transfer in which the MAE is lower. For both group combinations in Dagstuhl with the lowest IAAs (E,N and N,C), the Wasserstein distances are consistently lower than for E,C. This indicates systematic patterns of disagreement present to a different extent between groups.

In sum, we find considerable annotation variation within and across groups (RQ1), which causes limited transferability between groups (RQ2). 

\section{Discussion}
\label{sec:discussion}

\paragraph{Imbalance of AQ Datasets}
We identified a substantial number of datasets, covering nearly the entire taxonomy of quality categories.
The two dialectically-driven dimensions (reasonableness and deliberative norms) are less represented, likely due to the reduced focus on dialogical argumentation in CA and the more recent attention to deliberation. Unfortunately but not unexpectedly, we found a very uneven representation of language. Together with the sparsity of languages comes a lack of cultural diversity. As arguments are perceived differently across cultures \cite{han1994culture, Shen2023culture}, this gap should be closed.

\paragraph{Potential Bias in Annotator Representation}
The analysis of meta-information about annotators revealed that current datasets provide little and mostly fragmentary documentation about whose perception is captured. This lack of transparency raises the question of whether existing AQ datasets facilitate models that are biased to certain populations. We argue that certain characteristics of the annotators should be openly communicated, such as demographics (while adhering to privacy regulations); providing such data statements has been long called for by \citet{bender-friedman-2018-data}, but we recognize rare application in the field of AQ.

A prime example is gender, rarely specified in the reviewed datasets. More than half of the manually annotated datasets stem from the highly homogeneous group of students and experienced researchers with a background in computational linguistics or computer science (i.e., the in-house annotators). Given that these fields remain heavily male-dominated \cite{schluter-2018-glass}, such a selection strategy may unintentionally amplify gender bias. Equal considerations apply to other demographic attributes such as education and age. We also critically point to the practice of excluding annotators for inconsistency with their peers, as it wrongly assumes that diversity is per se an error and reinforces the formation of overly homogeneous groups.

Documenting such properties more transparently goes hand in hand with a raised awareness of the impact that annotator selection can have on the whole process. This is equally important for both the perspectivist turn and the established approach of aggregated ground truth (e.g., simple majority vote can exclude underrepresented groups entirely).

\paragraph{Potentials and Challenges for the Perspectivist Turn}

We identified a handful of non-aggregated AQ datasets to be suitable for enabling perspectivist model development, whereas many other datasets are insufficient in terms of a controlled selection of annotators, the number of annotators (in total and per item), and the dataset size.

As our experiments emphasize, dataset annotation decides who is represented by the model's output. We thus deem the collection of labeling decisions from annotators representative of a specific population crucial to building reasonable models for a desired target population. Likewise, the availability of individual-level annotator attributes is of high relevance as it facilitates modeling annotators more accurately, but also the development of perspectivist models at the group level.

It is furthermore vital to reflect on the number of annotators needed to build robust models. While 2--3 annotators may seem inadequate, there is no clear threshold for an appropriate group size. This also depends on the modeling goal; whether to develop perspectivist models in the literal sense, or to apply them for other purposes, such as personalization. Additionally, the number of dataset items and their coverage per annotator is essential. If items are sparsely labeled, the dataset may not provide enough information for individual annotators, an issue that has been raised by \citet{davani-etal-2022-dealing}.

In connection to the previous point of discussion, it is also crucial to keep in mind the risk of capturing spurious correlations between annotators' backgrounds and patterns in AQ assessment. This is especially important when working with datasets that were not explicitly created for perspectivist usage (e.g., in SIGIR-19). The question of how many data points per group are needed for the robust generalization of findings is an empirical one, usually constrained by budget limits. In such cases, the consideration of linguistic patterns alongside the task phenomenon may help support results.
\section{Conclusion}
\label{sec:conclusion}

A critical first step in developing perspectivist models for AQ assessment is suitable data. We identify several datasets as a learning ground for selected AQ categories in a non-aggregated way, while noting current shortcomings. Future datasets should (1)~{cover a diverse set of perspectives with respect to a reference population}, (2)~{collect annotator-specific attributes}, and (3)~{maintain an adequate size of total and individually labeled items}. The listed desiderata indicate that the perspectivist turn in AQ assessment requires resources to be invested in annotation quality and quantity. In this context, recent research on the potential of active learning methods for subjective NLP tasks may be relevant \cite{wang-plank-2023-actor,van-der-meer-etal-2024-annotator}.

Filling the resource gap will unlock the potential of the perspectivist turn in AQ assessment advocated in this paper. With suitable resources, future work can start leveraging the machine learning toolbox developed thus far in the perspectivist community (e.g., \citealp{davani-etal-2022-dealing,casola-etal-2023-confidence}).

Our extensive meta-information database will facilitate AQ research in general. One example is instruction-following LLMs, in which access to annotation guidelines is crucial \cite{wachsmuth-etal-2024-argument}. We make a significant contribution to this by expanding the number of publicly available guidelines,  provided as a central listing in our database.

\section*{Limitations}

\paragraph{Scope of the Dataset Search and Number of Annotators}
While we conducted a comprehensive and systematic search for datasets, we acknowledge that further datasets may be viewed as relevant that we did not cover. For example, the concepts of offensive or toxic language overlap with uncivil communication \cite{pachinger-etal-2023-toward}; or fallacies in propaganda detection. Moreover, while we believe that the validation on the 10\% sub-sample demonstrates the reliability of AQ annotations, carried out by a single reviewer on the remaining datasets, we cannot exclude the impact of subjectivity and potential errors. For both cases (scope of the datasets collected and potential categorization errors), however, we believe that the fact that the collection will be publicly available in form of a website will allow authors to reach out to us for updates.

\paragraph{Assumed Fit of the Selected Taxonomies to the Whole Dataset Collection} To categorize existing datasets, we have selected and applied two specific taxonomies \cite{wachsmuth-etal-2017-computational, friess2015deliberation}, one for argument quality and one for deliberation quality. While our motivations for this choice are strong, as discussed in Section \ref{sec:relatedwork}, we cannot in principle exclude that a different categorization would have had a better fit to the papers we collected. We believe, however, that our categorization approach, strictly based on consulting descriptions provided in papers and annotation guidelines (and contacting the authors directly when guidelines were not available) alleviates this limitation.

\paragraph{Disagreements: Between Valid Human Label Variation and Annotation Errors}
Disagreement among annotators can arise from various factors, among them subjectivity, but also annotation errors and ambiguity in the items to be labeled. Working with non-aggregated datasets thus always comes with the question of annotation reliability and how to distinguish potential annotation errors from valid label variations. First approaches are being developed to eventually complement perspectivist machine learning workflows \cite{weber-genzel-etal-2024-varierr}. Disentangling the different types of disagreements can improve data quality, which would not only benefit the perspectivist turn, but also the well-established approach of aggregated ground-truth. In both cases, the release of non-aggregated annotations is crucial.

While our pilot experiments in Section \ref{sec:pilot} exemplify the general consequences of pronounced differences in annotations across and within annotator groups, in this paper we do not specifically tackle the question of how much of these differences can be attributed to annotation error and how much to legitimately varied, subjective labeling decisions. However, the low to moderate IAA, even among the Dagstuhl experts --- who themselves developed the underlying taxonomy --- clearly indicates a significant degree of subjectivity in AQ perception.
\section*{Ethics Statement}

This paper shares the inherent ethical concerns raised by argument mining in general, and by the assessment of AQ in particular. The first concern is dual use of NLP tools developed to assess / generate persuasive arguments, which could be then employed to manipulate the public opinion. The second concern is bias that such tools may contain, leading to assess the higher quality of arguments based on spurious cues in the data.

Additionally, while perspectivism advocates for the inclusion of as many perspectives as possible, it inevitably calls for a data-greedy approach, that is, the more annotators, the better. This may come with a human cost:  it is typically achieved by crowd-sourcing, which is known to raise concerns about fair pay and treatment of the annotators. Also, the need of such expensive data collections may give an unfair advantage to highly funded researchers. Finally, the finer-grained the information about annotators is, the higher the privacy risks they are exposed to. Ideally, large-scale dataset surveys that aim at making resources aligned and comparable can enable data-greedy modeling without the need to annotate anew. 

Generally, annotator attributes must be collected in compliance with privacy regulations and with the consent of participants. Some of the datasets we reviewed were parsed from existing internet resources, with ground truth labels derived from platform users. We would like to emphasize that using any information for profiling users, especially when it may contain personally identifiable content, risks privacy violations and may raise ethical concerns.
\section*{Acknowledgments}

We would like to thank Ana Maria Lisboa dos Santos Cotovio for her support in the collection of the dataset and for setting up the website.

\bibliography{anthology,custom}
\bibliographystyle{acl_natbib}

\appendix
\onecolumn
\newpage
\twocolumn

\section{Details of the Dataset Search}
\label{sec:review-details}

\paragraph{Search Process}
Using Google site search, we identified 238 candidate papers. Based on a review of the titles, abstracts, and a skim of the content, we excluded papers that neither introduced novel data nor addressed AQ. We filtered duplicates --- papers that introduce the same dataset --- and kept the first published paper in such cases.

Datasets come in various formats, such as machine learning corpora, user studies, or manual content analyses. Our focus is on annotated datasets initially created as resources for machine learning. We also collect extensions of these datasets, even if introduced primarily for annotation studies (i.e., extensions of the Dagstuhl-ArgQuality corpus). Additionally, we include datasets originally developed for other contexts, such as qualitative content analysis in the social sciences, if they have later been used to train machine learning models. We collect datasets that were purposefully annotated through coordinated efforts in the context of scientific work, as well as such that have been crawled from existing sources that provide some sort of annotation (e.g., online debate portals).

\paragraph{Relevance}
For a dataset (and corresponding paper) to be considered relevant, it must encode at least one category (either major or sub-category) according to the taxonomy introduced in Section \ref{sec:review}. We do not require the entire dataset to focus exclusively on AQ assessment; it is sufficient if certain dimensions of the annotation code AQ.

Some works we reviewed consider topical relevance as part of AQ, either in terms of a premise’s relevance to a conclusion (as local relevance) or an argument’s relevance to the issue at hand (as global relevance). According to the definitions we follow, local relevance focuses on ``the contribution towards the acceptance or rejection of the argument's conclusion'', while global relevance addresses the ``contribution towards the issue's resolution''. However, it is debatable whether topical relevance consistently aligns with these goals. As briefly discussed by \citet{potthast2019argument}, although the notions of relevance in information retrieval and global relevance share similarities, the authors chose to keep the concepts distinct. In our review, we similarly excluded such cases.

\paragraph{Reddit CMV (ChangeMyView)}
Several works use the Reddit CMV subreddit as a source for pre-annotated data on persuasion. Due to the at times significant overlap between these datasets, in case of duplicate content we focus on the CMV crawl that was introduced first. We include additional datasets in our database only if they (a) offer distinct value by including more recent data from the subreddit, or (b) provide further annotations of AQ for the same time span.

\paragraph{Dataset Extensions}
Nine datasets extend other AQ datasets directly by adding further samples (4), by re-annotating the same samples (3), by adding a further modality (1) or by releasing the non-aggregated version of a previously aggregated dataset release (1).

\section{Collection of Properties in the Database}
\label{sec:database-details}

We publish the results of our systematic literature review in the form of an online database to inform future research in AQ. This database solely contains meta-information on the identified datasets. Each dataset is represented as a row, while the columns contain a comprehensive set of characteristics describing the datasets.
First, we collect basic information about each dataset, namely the \textbf{dataset name}, whether the dataset is an \textbf{extension of another dataset} (\textit{extension of samples}, \textit{extension of annotators}, \textit{multimodal extension}, and \textit{non-aggregated version}), the \textbf{availability of the dataset} (\textit{online}, \textit{upon request}, \textit{not publicly available}, \textit{no indication}), --- if available online --- the \textbf{link to the resource} and \textbf{status of the link} (\textit{accessible} or \textit{not accessible}; as of August 2024), the \textbf{license}, whether the data contains \textbf{manual annotation}\footnote{We differentiate between datasets produced through coordinated manual annotation studies and those extracted from existing resources, like debate forums, where labels for certain AQ phenomena are inherently present.}, and --- in case of manual annotation --- the \textbf{availability of annotation guidelines} (\textit{online} (with reference to location), \textit{upon request}, \textit{not available in full}, \textit{not publicly available}, and \textit{no indication}).
Corresponding information on the paper that introduced the dataset is provided through \textbf{paper title}, the \textbf{paper authors}, the \textbf{year} of publication, and the \textbf{paper URL}, together with the targeted \textbf{research community} (based on the publication venue).

Going more into detail on the single datasets, we indicate the \textbf{size} of the dataset in terms of \textbf{units of annotation}, the \textbf{genre}, the \textbf{modality} (either \textit{text} or \textit{multimodal}), and the \textbf{language} represented.
In line with our goal of identifying AQ datasets, we provide a \textbf{textual description of AQ categories} contained as described in the words of the respective paper and indicate via \textbf{AQ taxonomy codes} (see Table \ref{tab:taxonomy}) which categories of the taxonomy the dataset covers.
For manual annotation datasets, we track the \textbf{aggregation method} that was used to form an aggregated ground truth (in case of aggregated ground truth datasets), the inter-annotator-agreement through \textbf{IAA score} and \textbf{IAA measure}, and whether there is a \textbf{non-aggregated version} available (\textit{no}, \textit{yes (annotator-specific)}, \textit{yes (distribution-based)}, and \textit{no indication}).

For exploring whose perspectives (in terms of perceiving AQ) are represented in current datasets, we gathered details on the \textbf{selection method of annotators} from the authors' descriptions, whether they are \textbf{in-house or crowd} workers, whether they are \textbf{expert or novice} annotators, and any available \textbf{annotator attributes} provided in the associated paper or dataset. Additionally, we account for the \textbf{number of annotators per item} and the \textbf{number of annotators in total}.

We also examined the authors of the arguments, referred to as ``argument producers''. For these producers, we give the number of distinct individuals represented in the dataset (\textbf{number of producers}) and reviewed the available socio-demographic information provided in the corresponding papers or datasets (\textbf{producer attributes}).

\section{Details of the Non-Aggregated Datasets}
\label{sec:description-non-aggr-datasets}

\paragraph{CrowDEA Ideas dataset \cite{Baba_Li_Kashima_2020}}
This Japanese-language corpus contains preference labels for solution proposals to everyday life questions. A total of 16k argument pairs were annotated by 20 different workers, drawn from a pool of 257 crowd workers (of which no further information is provided).

\paragraph{CIMT PartEval Argument Concreteness Corpus \cite{romberg-etal-2022-corpus}}
As a tool to support public institutions in Germany in evaluating citizen contributions, this dataset provides annotations on how concrete an argument is introduced. Five student annotators fully labeled a total of 1127 argument units. Acknowledging the subjectivity of the task, the dataset was explictely published in a non-aggregated way.

\paragraph{TYPIC \cite{naito-etal-2022-typic}}
To offer feedback on flaws in Japanese students' arguments, the authors took an approach that first provides diagnostic comments describing weaknesses in the arguments. These comments are then mapped to AQ criteria: local acceptability, local sufficiency, local relevance, global relevance, and global sufficiency. Four experts generated 1,082 diagnostic comments for 197 Japanese-language arguments, with each argument receiving two labels. The categorization was conducted by one to two annotators.

\paragraph{Argument Validity and Novelty Prediction Shared Task \cite{heinisch-etal-2023-architectural}}
The non-aggregated version of the dataset from the Argument Validity and Novelty Prediction Shared Task \cite{heinisch-etal-2022-overview}, co-located with the ArgMining workshop 2022, was published in later work for use in multi-annotator models. 1474 premise-conclusion pairs from English-language online debate portals come with three annotations per item, drawing from a pool of five student experts.

\paragraph{MAFALDA \cite{helwe-etal-2024-mafalda}}
The authors developed a hierarchical taxonomy of fallacies, resulting in the MAFALDA corpus, which contains 268 argumentative spans drawn from English-language news articles, social media, and political debates. Four of the authors annotated the spans, with each item receiving up to four labels.

\paragraph{UMOD dataset (User Moderation in Online Discussions) \cite{falk-etal-2024-moderation}}
The study focuses on annotating characteristics of user-driven moderation in online discussions, among them the constructiveness of such comments. The dataset, sourced from English Reddit's Change My View (CMV), contains 1,000 comment-reply pairs. Each pair was annotated by nine crowd workers, with a total of 90 workers participating. To provide a more nuanced understanding of the annotators, socio-demographic information including race, sex, age, and role) are collected too.


\paragraph{The Persuasion and Personality Corpus \cite{lukin-etal-2017-argument}}
An explicitly perspectivist corpus was introduced to investigate differences in the perception of argument effectiveness. To this end, for 637 crowd workers --- representative of the U.S population --- stance changes elicited by presented social media arguments were recorded. At the same time, Big Five personality traits were collected, alongside further socio-demographic information (age, gender, political view, education, civic engagement, religion, spirituality, employment status, and income). The resulting dataset contains 100 items, each annotated by 20 workers.

\paragraph{Webis-Editorial-Quality-18 corpus \cite{el-baff-etal-2018-challenge}}
This corpus was created for assessing AQ of news editorials. 
Resembling \citet{lukin-etal-2017-argument}, different perceptions were considered using the Big Five personality traits and the political leaning. 1000 news editorials were annotated by three liberals and three conservatives each, 24 crowd workers from the U.S. in total.


\paragraph{n.a. \cite{hunter2017empirical}}
To study the personalization of argumentation, 50 crowd workers (no further details about the workers were provided) annotated a set of 30 English-language arguments for believability, convincingness, and appeal.

\paragraph{SIGIR-19 \cite{potthast2019argument}}
A resource that codes the logical, rhetorical, and dialectical quality of arguments in the context of information retrieval. 40 student volunteers annotated 494 online debate portal arguments in English, with one annotator per item. While gender and age are indicated annotator-specific, from the paper we learn about the political leaning (80\% vote for left-wing, green parties) which may impact the perspectives of annotators.

\paragraph{argumentation attitude dataset \cite{brenneis-etal-2021-will}}
The German-language dataset was collected from a deliberation platform for political opinion-forming. A total of 946 arguments were rated in four waves based on individual conviction and argument strength. The 674 crowd workers constitute a representative selection of the German online population in terms of age, gender and education. Each argument received ratings from one to 147 individuals.
A unique feature of this dataset is that it captures not only the perception of argument effectiveness but also the writing style of the individual members of the crowd, as they contribute part of the arguments as well.

\paragraph{Dagstuhl-15512-ArgQuality \cite{wachsmuth-etal-2017-computational}}
Along with introducing their taxonomy of AQ, the authors provide a dataset: 320 English-language arguments from online debate portals were rated across 15 categories by three expert annotators.
While there is no attribution to the annotator id, distributional information about gender, education, and employment is given, providing some information about the annotators' socio-demographics.

\paragraph{Dagstuhl-15512-ArgQuality: extension 1 \cite{wachsmuth-etal-2017-argumentation}}
The same dataset was annotated by 102 crowd workers (no further details about the workers were provided), with 10 workers assigned to each argument.

\paragraph{Dagstuhl-15512-ArgQuality: extension 2 \cite{mirzakhmedova2024reliable}}
The dataset was also annotated by 108 novice in-house annotators, all undergraduate students without prior experience in computational argumentation, with up to 10 raters per argument.
\textit{Note: The official release includes a subset of annotators only, which are not uniquely identifiable.}

\paragraph{GAQCorpus \cite{lauscher-etal-2020-rhetoric}}
The authors introduced one of the largest corpora for assessing multiple key aspects of AQ. The corpus includes 5,285 arguments from diverse domains (debate forums, question-answering forums, and review forums), annotated for cogency, effectiveness, reasonableness, and overall quality. Annotation was conducted using a mix of three expert annotators and 24 crowd workers. Some items were annotated exclusively by the crowd (10 per item), others exclusively by the experts (up to three annotations per item), and a portion of the dataset was jointly annotated by both groups (up to 13 annotations per item).
\textit{Note: The corpus contains varying numbers of annotators for different parts of the data.}

\paragraph{EuropolisAQ \cite{falk-lapesa-2022-scaling}}
The dataset contains 513 transcribed speeches from a transnational deliberative poll. It builds on the Europolis corpus \cite{Gerber_Baechtiger_Shikano_Reber_Rohr_2018}, labeled with deliberative norms, and extends it with annotations of the other dimensions of AQ, namely cogency, effectiveness, reasonableness, and overall AQ. Each item was annotated by one or both of two expert annotators. To the best of our knowledge, this is the only corpus that provides comprehensive coverage of the four major aspects of AQ according to our taxonomy --- while not all dimensions are available in non-aggregated format.

\paragraph{ArgQ! \cite{silva2021quality}}
The authors adapt the rhetorical effectiveness dimension of \citet{wachsmuth-etal-2017-computational}'s taxonomy for use with Twitter data and expert annotate 352 argumentative tweets from the Brazilian political context accordingly. Each argument contains four labels from a total of four different annotators.

\paragraph{UKPConvArg1 \cite{habernal-gurevych-2016-argument}}
The dataset consists of 16k pairwise comparisons of arguments from online debate portals with respect to their convincingness.
Each pair was annotated by five crowd workers, with about 3,900 workers participating, all from the U.S. Additionally, the workers' stance towards the discussed topic was tracked, as it could likely influence their perspective during the assessment.

\paragraph{UKPConvArg2 \cite{habernal-gurevych-2016-makes}}
In addition to the overall assessment of effectiveness in UKPConvArg1, this dataset provides a categorization into finer attributes based on textual decision rationales formulated by the individual workers in the previous annotation. The attributes give reasons for why an annotator found an argument to be convincing, covering primarily logical and rhetorical categories of the taxonomy. Each of the 70k reason units comes with 5 annotations from a pool of 776 crowd workers.

\paragraph{Essay Argument Organization Dataset \cite{persing-etal-2010-modeling}}
Targeting the automatic evaluation of persuasive essays, this study addresses the evaluation dimension of organization, which assesses how well an essay is structured to logically develop an argument.
1003 persuasive essays, written by a diverse set of English learners from 15 native languages, were rated by one to six annotators.
\textit{Note, that in this datasets there is only a distribution of annotations given per item, no attribution to unique annotator.}

\paragraph{Appropriateness Corpus \cite{ziegenbein-etal-2023-modeling}}
The authors introduce a refined definition of the subcategory ``appropriateness'' by offering a more sophisticated interpretation. Using 2,191 arguments from existing AQ corpora, they had appropriateness annotated by three crowd workers for each argument. While there is no attribution to the annotator id, distributional information about gender and mother tongue is given, providing some information about the annotators' socio-demographics.

\paragraph{UKP-InsufficientArguments \cite{stab-gurevych-2017-recognizing}}
The authors present a corpus of persuasive essays annotated for local sufficiency. For inter-annotator-agreement, 433 of the 1029 arguments were annotated by three expert annotators.

\paragraph{Webis-ArgRank-17 Dataset \cite{wachsmuth-etal-2017-pagerank}}
This dataset is on the global relevance of arguments. Mainly created automatically, it contains a manually annotated ground truth of 110 arguments annotated by seven experts from computational linguistics and information retrieval each.

\paragraph{StoryARG \cite{falk-lapesa-2023-storyarg}}
The authors developed a corpus focused on narratives and personal experiences in argumentative texts. Each argument (2451 in total) was annotated by one to four annotators, with a total of four annotators participating in the project. Distributional information on gender, education, and country of origin is provided for the annotators.

\section{Definition of Argument Quality}
\label{app:taxonomy}

\begin{table*}[ht]
\small\renewcommand{\arraystretch}{1.15}
\setlength{\tabcolsep}{2.5pt}
\centering
\begin{tabular*}{\linewidth}{p{0.20\linewidth}p{0.78\linewidth}}
\toprule 
\textbf{Category} & \textbf{Description}\\
\midrule
\bf Logical cogency & An argument is cogent if it has acceptable premises that are relevant to its conclusion and that are sufficient to draw the conclusion.\\
Local acceptability&A premise of an argument is acceptable if it is rationally worthy of being believed to be true.\\
Local relevance&A premise of an argument is relevant if it contributes to the acceptance or rejection of the argument's conclusion.\\
Local sufficiency&An argument's premises are sufficient if, together, they give enough support to make it rational to draw its conclusion.\\
\midrule
\bf Rhetorical effectiveness &Argumentation is effective if it persuades the target audience of (or corroborates agreement with) the author’s stance on the issue.\\
Clarity&Argumentation has a clear style if it uses correct and widely unambiguous language as well as if it avoids unnecessary complexity and deviation from the issue.\\
Credibility&Argumentation creates credibility if it conveys arguments and similar in a way that makes the author worthy of credence.\\
Appropriateness&Argumentation has an appropriate style if the used language supports the creation of credibility and emotions as well as if it is proportional to the issue.\\
Emotional appeal&Argumentation makes a successful emotional appeal if it creates emotions in a way that makes the target audience more open to the author’s arguments.\\
Arrangement&Argumentation is arranged properly if it presents the issue, the arguments, and its conclusion in the right order.\\
\midrule
\bf Dialectical reasonableness&Argumentation is reasonable if it contributes to the issue’s resolution in a sufficient way that is acceptable to the target audience.\\
 Global acceptability&Argumentation is acceptable if the target audience accepts both the consideration of the stated arguments for the issue and the way they are stated.\\
 Global relevance&Argumentation is relevant if it contributes to the issue’s resolution, i.e., if it states arguments or other information that help to arrive at an ultimate conclusion.\\
 Global sufficiency&Argumentation is sufficient if it adequately rebuts those counterarguments to it that can be anticipated.\\
\midrule
\bf Deliberative norms& Argumentation adheres to deliberative norms if it promotes a respectful and inclusive exchange of rational or alternative forms of argument, with the aim of reaching mutual understanding.\\
Rationality&Deliberation is rational if it is centered on arguments that are supported by solid evidence (either through facts that can be verified or through a shared understanding of moral or normative behavior), arguments and further information that are put forward in the discourse are relevant to the topic, and an informed ground for discussion is built (e.g., through providing an information base in the beginning of the discussion, or information requests by participants to make the discourse more informed). With respect to the dimensions of argumentation quality, the focus is on normatively well-reasoned arguments and not on how good these are perceived by the target audience.\\
 Interactivity&Deliberation is interactive if the participants actively engage with each other by exchanging arguments in a way where they listen to the other participants, understand their perspective, and relate to it in a substantive way (e.g., by valuing, critiquing, or countering other's arguments, or question asking).\\
 Equality&Deliberation is equal if all participants (irrespective of their background) have the same opportunity to participate by putting forward their own arguments and responding to other's claims. This dimension of deliberation quality tackles inclusiveness and accessibility.\\
 Civility&Deliberation is civil if the participants show respect to the other participants by recognizing them as equal actors in the discourse and acknowledging the value of opposing claims. Respectful interaction is regarded as a prerequisite for participants to be convincable by other opinions and to reach a consensus decision in the sense of deliberation.\\
 Common good reference&Deliberation is oriented towards the common good if arguments are justified by promoting the well-being of the community as a whole rather than serving the interests of narrow interest groups. What exactly is considered the common good can include different goals, such as achieving the best outcome for the greatest number of people or prioritizing the needs of the most disadvantaged members of society. The joint focus on a common good is regarded as a basis for participants with diverse interests to be able to convince each other.\\
 Constructiveness&Deliberation is constructive if it contributes to finding a consensus decision for the issue of discussion through actions like proposing new solutions, searching for common ground, appeals for mobilisation, or questions addressed to the community.\\
 Alternative forms of communication&In scenarios in which not all participants are able to adhere to the rigid concept of rational argumentation based on verifiable facts, other forms of communication can provide a valuable resource for good deliberation. These include storytelling, testimonies, narratives, emotional talk, casual talk, humor, or even gossip.\\
\midrule
\bf Overall quality&An overarching measure of the quality of arguments.\\
\bottomrule
\end{tabular*}
\caption{Taxonomy of argument quality. The definitions of the first three dimensions are taken verbatim from \citet{wachsmuth-etal-2017-computational}. The definitions of the last dimension are based on \citet{friess2015deliberation}.}
\label{tab:taxonomy} 
\end{table*}

Table~\ref{tab:taxonomy} shows the original definitions of all AQ dimensions considered in this work.

\section{List of AQ Datasets}
\label{sec:relevant-datasets}

Table~\ref{tab:datasets} lists the 103 datasets for AQ that we identified in our systematic literature search. 

\onecolumn

{\footnotesize
\begin{longtable}{P{4.1cm}P{1.9cm}P{3cm}P{1cm}p{4cm}}
\toprule 
\textbf{Dataset Name} & \textbf{Extension of Previous} & \textbf{Paper} & \textbf{Community} & \textbf{AQ}\\ 
\midrule
\endfirsthead
\multicolumn{5}{c}%
{{\bfseries \tablename\ \thetable{} -- continued from previous page}} \\
\midrule 
\textbf{Dataset Name} & \textbf{Extension of Previous} & \textbf{Paper} & \textbf{Community} & \textbf{AQ}\\ \hline 
\endhead
\midrule 
\multicolumn{5}{r}{{Continued on next page}} \\ 
\midrule
\endfoot
\bottomrule \\[-2.5ex]
\caption{Overview of AQ datasets, including the name of the dataset (if provided, otherwise n.a.), whether and how it extends the previously listed dataset, the publication in which the dataset was introduced, the research community targeted (NLP, AI: artificial intelligence, CA: computational argumentation, CS: computer science, HCI: human computer interaction, IR: information retrieval, SocSci: Social Sciences, Web) and the assigned categories of AQ.}
\label{tab:datasets} 
\endlastfoot
\textbf{Essay Argument Organization Dataset} && \citet{persing-etal-2010-modeling} & NLP & Arrangement\\
\textbf{n.a.} && \citet{cabrio-villata-2012-combining} & NLP & Global acceptability\\
\textbf{Essay Thesis Clarity Dataset} && \citet{persing-ng-2013-modeling} & NLP & Clarity\\
\textbf{n.a.} && \citet{xu2014identifying} & AI & Deliberative norms\\
\textbf{Essay Prompt Adherence Dataset} && \citet{persing-ng-2014-modeling} & NLP & Clarity\\
n.a. && \citet{coe2014online} & SocSci & Civility\\
\textbf{Essay Argument Strength Dataset} && \citet{persing-ng-2015-modeling} & NLP & Effectiveness\\
\textbf{Intelligence Squared Debates Corpus} && \citet{zhang-etal-2016-conversational} & NLP & Effectiveness\\
\textbf{n.a.} && \citet{niculae-danescu-niculescu-mizil-2016-conversational} & NLP & Constructiveness\\
\textbf{n.a.} && \citet{braunstain2016supporting} & IR & Local Relevance\\
\textbf{UKPConvArg1} && \citet{habernal-gurevych-2016-argument} & NLP & Effectiveness\\
\textbf{UKPConvArg2} && \citet{habernal-gurevych-2016-makes} & NLP & Local relevance, local sufficiency, clarity, credibility, appropriateness, emotional appeal, overall quality\\
\textbf{CMV} && \citet{tan2016winning} & Web & Effectiveness\\
\textbf{Webis-ArgRank-17 Dataset} && \citet{wachsmuth-etal-2017-pagerank} & NLP & Global relevance\\
\textbf{n.a.} && \citet{habernal-etal-2017-argotario} & NLP & Local sufficiency, clarity, appropriateness, global sufficiency, global relevance\\
n.a. & more samples & \citet{habernal-etal-2018-adapting} & NLP & Local sufficiency, clarity, appropriateness, global sufficiency, global relevance\\
\textbf{The Persuasion and Personality Corpus} && \citet{lukin-etal-2017-argument} & NLP & Effectiveness\\
\textbf{Dagstuhl-15512-ArgQuality} && \citet{wachsmuth-etal-2017-computational} & NLP & Cogency, local acceptability, local relevance, local sufficiency, effectiveness, clarity, credibility, appropriateness, emotional appeal, arrangement, reasonableness, global acceptability, global relevance, global sufficiency, overall quality\\
n.a. & more annotators & \citet{wachsmuth-etal-2017-argumentation} & NLP & Cogency, local acceptability, local relevance, local sufficiency, effectiveness, clarity, credibility, appropriateness, emotional appeal, arrangement, reasonableness, global acceptability, global relevance, global sufficiency, overall quality\\
n.a. & more annotators & \citet{mirzakhmedova2024reliable} & CA & Cogency, local acceptability, local relevance, local sufficiency, effectiveness, clarity, credibility, appropriateness, emotional appeal, arrangement, reasonableness, global acceptability, global relevance, global sufficiency, overall quality\\
\textbf{n.a.} && \citet{beigman-klebanov-etal-2017-detecting} & NLP & Overall quality\\
\textbf{n.a.} && \citet{hunter2017empirical} & AI & Effectiveness\\
\textbf{YNACC} && \citet{napoles-etal-2017-finding} & NLP & Overall quality\\
\textbf{UKP-InsufficientArguments} && \citet{stab-gurevych-2017-recognizing} & NLP & Local sufficiency\\
\textbf{Debate Argument Persuasiveness Data} && \citet{persing2017convince} & AI & Effectiveness\\
\textbf{CDCP} && \citet{park-cardie-2018-corpus} & NLP & Clarity\\
\textbf{n.a.} && \citet{habernal-etal-2018-name} & NLP & Appropriateness, global sufficiency\\
\textbf{Webis-Editorial-Quality-18} && \citet{el-baff-etal-2018-challenge} & NLP & Global acceptability\\
\textbf{CGA-WIKI} && \citet{zhang-etal-2018-conversations} & NLP & Appropriateness, civility\\
n.a. & more samples & \citet{chang-danescu-niculescu-mizil-2019-trouble} & NLP & Appropriateness, civility\\
\textbf{n.a.} && \citet{Gerber_Baechtiger_Shikano_Reber_Rohr_2018} & SocSci & Rationality, interactivity, civility, common good reference, alternative forms of communication\\
\textbf{Essay Argument Persuasiveness Dataset} && \citet{carlile-etal-2018-give} & NLP & Effectiveness\\
\textbf{Webis-WikiDebate-18} && \citet{al-khatib-etal-2018-modeling} &  NLP & Rationality, constructiveness\\
\textbf{DDO} && \citet{durmus-cardie-2019-corpus} & NLP & Effectiveness\\
\textbf{n.a.} && \citet{yang2019corpus} & NLP & Local acceptability\\
n.a. & more annotators & \citet{yang-etal-2019-nonsense} & NLP & Local acceptability\\
\textbf{IBM-EviConv} && \citet{gleize-etal-2019-convinced} & NLP & Effectiveness\\
\textbf{SIGIR-19} && \citet{potthast2019argument} & IR & Cogency, effectiveness, reasonableness\\
\textbf{IBM-ArgQ-Args} && \citet{toledo-etal-2019-automatic} & NLP & Overall quality\\
\textbf{IBM-ArgQ-Pairs} && \citet{toledo-etal-2019-automatic} & NLP & Overall quality\\
\textbf{Essay Thesis Strength Dataset} && \citet{ke-etal-2019-give} & NLP & Effectiveness\\
\textbf{n.a.} && \citet{potash-etal-2019-ranking} & NLP & Effectiveness\\
\textbf{n.a.} && \citet{durmus-etal-2019-role} & NLP & Global relevance\\
\textbf{CGA-CMV} && \citet{chang-danescu-niculescu-mizil-2019-trouble} & NLP & Appropriateness, civility\\
\textbf{n.a.} && \citet{atkinson-etal-2019-gets} & NLP & Effectiveness\\
\textbf{IBM-ArgQ-Rank-30kArgs} && \citet{Gretz_Friedman_Cohen-Karlik_Toledo_Lahav_Aharonov_Slonim_2020} & AI & Overall quality\\
\textbf{CrowDEA Ideas} && \citet{Baba_Li_Kashima_2020} & HCI & Overall quality\\
\textbf{n.a.} && \citet{jo-etal-2020-detecting} & NLP & Global sufficiency\\
\textbf{Webis-ArgQuality-20} && \citet{gienapp-etal-2020-efficient} & NLP & Cogency, effectiveness, reasonableness\\
\textbf{Webis-CMV-20} && \citet{al-khatib-etal-2020-exploiting} & NLP & Effectiveness\\
\textbf{Chinese-Essay-Dataset-For-Organization-Evaluation} && \citet{ijcai2020p536} & AI & Arrangement\\
\textbf{XArgMining Dataset ArgsHG} && \citet{toledo-ronen-etal-2020-multilingual} & NLP & Overall quality\\
\textbf{n.a.} && \citet{dumani2020ranking} & IR & Cogency, effectiveness, reasonableness\\
\textbf{GAQCorpus} && \citet{lauscher-etal-2020-rhetoric} & NLP & Cogency, effectiveness, reasonableness, overall quality\\
\textbf{n.a.} && \citet{sahai-etal-2021-breaking} & NLP & Local acceptability, local relevance, local sufficiency, clarity, global relevance\\
\textbf{argumentation-attitude-dataset} && \citet{brenneis-etal-2021-will} & NLP & Effectiveness\\
\textbf{WikiDisputes} && \citet{de-kock-vlachos-2021-beg} & NLP & Constructiveness\\
\textbf{GermEval 2021} && \citet{risch-etal-2021-overview} & NLP & Rationality, interactivity, civility\\
\textbf{Touché21-Argument-Retrieval-for-Controversial-Questions} && \citet{bondarenko2021touche} & IR & Effectiveness\\
\textbf{Touché21-Argument-Retrieval-for-Comparative-Questions} && \citet{bondarenko2021touche} & IR & Effectiveness\\
\textbf{ArgQ!} && \citet{silva2021quality} & NLP & Effectiveness, clarity, credibility, emotional appeal, arrangement\\
\textbf{\#meinfernsehen2021} && \citet{2022:gerlach:meinferns} & SocSci & Rationality, interactivity, civility, alternative forms of communication\\
\textbf{CIMT PartEval Argument Concreteness Corpus} && \citet{romberg-etal-2022-corpus} & NLP & Clarity\\
\textbf{Advocacy Campaign Corpus} && \citet{kornilova-etal-2022-item} & NLP & Effectiveness\\
\textbf{Webis-Persuasive-Debaters-on-Reddit-CMV-2022} && \citet{wiegmann-etal-2022-analyzing} &  NLP & Effectiveness\\
\textbf{AM2} && \citet{chen-etal-2022-argument} & NLP & Effectiveness\\
\textbf{n.a.} && \citet{musi2022fake} & SocSci & Local relevance, local sufficiency, clarity, credibility, global relevance\\
\textbf{KODIE} && \citet{heinbach2022effects} & SocSci & Rationality, interactivity, civility, alternative forms of communication\\
\textbf{ElecDeb60To16-fallacy} & & \citet{ijcai2022p575} & AI & Local acceptance, local sufficiency, appropriateness, global sufficiency\\
ElecDeb60to20 & more samples & \citet{goffredo-etal-2023-argument} & NLP & Local acceptance, local sufficiency, appropriateness, global sufficiency\\
MM-USED-fallacy & multimodality & \citet{mancini-etal-2024-multimodal} & NLP & Local acceptance, local sufficiency, appropriateness, global sufficiency\\
\textbf{Persuasive Essays - Argument Quality Dataset} && \citet{marro-etal-2022-graph} & NLP & Cogency, effectiveness, reasonableness\\
\textbf{WikiTactics} && \citet{de-kock-vlachos-2022-disagree} & NLP & Constructiveness\\
\textbf{ImageArg} && \citet{liu-etal-2022-imagearg} & NLP & Effectiveness\\
ImageArg-Shared-Task & more samples & \citet{liu-etal-2023-overview} & NLP & Effectiveness\\
\textbf{n.a.} && \citet{2022:esau:kommunikat} & SocSci & Rationality, civility, constructiveness, alternative forms of communication\\
\textbf{Logic} && \citet{jin-etal-2022-logical} & NLP & Cogency, local acceptability, local relevance, local sufficiency, clarity, appropriateness, global relevance, global sufficiency\\
\textbf{LogicClimate} && \citet{jin-etal-2022-logical} & NLP & Cogency, local acceptability, local relevance, local sufficiency, clarity, appropriateness, global relevance, global sufficiency\\
\textbf{ABMPC} && \citet{wambsganss-niklaus-2022-modeling} & NLP & Effectiveness\\
\textbf{CLIMATE} && \citet{alhindi-etal-2022-multitask} & NLP & Local relevance, local sufficiency, clarity, credibility, global sufficiency, global relevance\\
\textbf{Argument Validity and Novelty Prediction Shared Task} && \citet{heinisch-etal-2022-overview} & NLP & Local sufficiency, global relevance\\
n.a. & non-aggregated & \citet{heinisch-etal-2023-architectural} & NLP & Local sufficiency, global relevance\\
\textbf{Touché22-Argument-Retrieval-for-Controversial-Questions} && \citet{bondarenko2022touche} & IR & Effectiveness\\
\textbf{Touché22-Argument-Retrieval-for-Comparative-Questions} && \citet{bondarenko2022touche} & IR & Effectiveness\\
\textbf{EuropolisAQ} && \citet{falk-lapesa-2022-scaling} & NLP & Cogency, effectiveness, reasonableness, overall quality\\
\textbf{TYPIC} && \citet{naito-etal-2022-typic} & NLP & local acceptability, local relevance, local sufficiency, clarity, global relevance, global sufficiency\\
\textbf{ArgAnalysis35K} && \citet{joshi-etal-2023-arganalysis35k} & NLP & Overall quality\\
\textbf{n.a.} && \citet{reveilhac2023comparing} & SocSci & Deliberative norms, rationality, interactivity, civility, constructiveness, alternative forms of communication\\
\textbf{DeliData} && \citet{karadzhov2023delidata} & HCI & Constructiveness\\
\textbf{CoRe} && \citet{salamat2023raise} & IR & Effectiveness\\
\textbf{Fallacies of Appeal to Emotions Corpus} && \citet{nieto-benitez2023fallacies} & CS & Local sufficiency\\
\textbf{Appropriateness Corpus} && \citet{ziegenbein-etal-2023-modeling} & NLP & Appropriateness\\
\textbf{Touché23-Argument-Retrieval-for-Controversial-Questions} && \citet{bondarenko2023touche} & IR & Effectiveness\\
\textbf{StoryARG} && \citet{falk-lapesa-2023-storyarg} & NLP & Alternative forms of communication\\
\textbf{FALLACIES} && \citet{hong-etal-2024-closer} & NLP & Local acceptability, local relevance, local sufficiency, clarity, appropriateness, credibility, emotional appeal, global relevance, global sufficiency\\
\textbf{DARIUS} && \citet{schaller-etal-2024-darius} & NLP & Local acceptability, local relevance, clarity\\
\textbf{ICLE++} && \citet{li-ng-2024-icle} & NLP & effectiveness, clarity, arrangement\\
\textbf{MAFALDA} && \citet{helwe-etal-2024-mafalda} & NLP & Local acceptability, local relevance, local sufficiency, clarity, appropriateness, global sufficiency, global relevance\\
\textbf{MISSCI} && \citet{glockner-etal-2024-missci} & NLP & Local acceptability, local relevance, local sufficiency, clarity, global relevance\\
\textbf{UMOD} && \citet{falk-etal-2024-moderation} & NLP & Constructiveness\\
\textbf{COVID-19 Discourse Corpus} && \citet{falk-lapesa-2024-stories} & NLP & Alternative forms of communication\\
\textbf{ArgSum Dataset} && \citet{li-etal-2024-side} & NLP & Effectiveness\\
\end{longtable}}

\end{document}